\pdfoutput=1
\documentclass[11pt]{article}

\usepackage{EMNLP2023}

\usepackage{times}
\usepackage{latexsym}

\usepackage[T1]{fontenc}
\usepackage[utf8]{inputenc}

\usepackage{microtype}

\usepackage{inconsolata}

\usepackage{url}            % simple URL typesetting
\usepackage{booktabs, multirow, tabularx}       % professional-quality tables
\usepackage{amsfonts}       % blackboard math symbols
\usepackage{nicefrac}       % compact symbols for 1/2, etc.
\usepackage{microtype}      % microtypography
\usepackage{graphicx}
\usepackage{natbib}
\usepackage{doi}
\usepackage{color}
\usepackage{amsmath}
\usepackage{mathtools, cuted}
\usepackage{etoolbox}
\usepackage{changepage}
\usepackage{pdflscape}
\usepackage{multicol}
\usepackage{balance}
\AfterEndEnvironment{strip}{\leavevmode}

\newcommand{\makecell}[2][@{}c@{}]{\begin{tabular}{#1}#2\end{tabular}}
\newcommand{\nce}{\mathrm{NCE}}

\newcommand{\JinaM}{\textsc{Jina Embeddings}}

\newcommand{\JinaS}{\href{https://huggingface.co/jinaai/jina-embedding-s-en-v1}{\texttt{jina-small-v1}}}
\newcommand{\JinaB}{\href{https://huggingface.co/jinaai/jina-embedding-b-en-v1}{\texttt{jina-base-v1}}}
\newcommand{\JinaL}{\href{https://huggingface.co/jinaai/jina-embedding-l-en-v1}{\texttt{jina-large-v1}}}

\title{\JinaM: A Novel Set of High-Performance Sentence Embedding Models}
\author{Michael G\"unther\and Louis Milliken\and Jonathan Geuter\\ {\bf Georgios Mastrapas}\and {\bf Bo Wang} \and {\bf Han Xiao} \\
\\
	Jina AI\\
	Ohlauer Str. 43, 10999 Berlin, Germany \\
	\texttt{\{michael.guenther,louis.milliken,jonathan.geuter,} \\
\texttt{georgios.mastrapas,bo.wang,han.xiao\}@jina.ai}
 }

\date{2023/07/20}

\hypersetup{
pdftitle={\JinaM: A Novel Set of High-Performance Sentence Embedding Models},
pdfauthor={Michael G\"unther, Louis Milliken, Jonathan Geuter, Georgios Mastrapas, Bo Wang, Han Xiao},
pdfkeywords={Embeddings, Large Language Models, Training},
}

\begin{document}
\maketitle
\begin{abstract}
\JinaM\ constitutes a set of high-performance sentence embedding models adept at translating textual inputs into numerical representations, capturing the semantics of the text.
These models excel in applications like dense retrieval and semantic textual similarity. This paper details the development of \JinaM, starting with the creation of high-quality pairwise and triplet datasets.
It underlines the crucial role of data cleaning in dataset preparation, offers in-depth insights into the model training process, and concludes with a comprehensive performance evaluation using the Massive Text Embedding Benchmark (MTEB).
Furthermore, to increase the model's awareness of grammatical negation, we construct a novel training and evaluation dataset of negated and non-negated statements, which we make publicly available to the community.
\end{abstract}

\section{Introduction}
\label{sec:introduction}
Sentence embedding models are an effective instrument for encoding the semantic nuances of words, phrases, and larger textual units into a continuous vector space. They encapsulate the complexities of contexts and lexical and grammatical interrelationships within a text, facilitating downstream tasks like information retrieval, semantic similarity evaluation, and text classification.

Despite the potential of these models, questions remain about the effectiveness of different data preprocessing strategies, the optimal loss function for training sentence embedding models, and the impact on performance of increasing the number of model parameters. This paper addresses these challenges.

We have develop a novel dataset specifically to train our sentence embedding models.
Furthermore, we design a dataset specifically to sensitize our models to distinguish negations of statements from confirming statements.
This paper also presents \textit{\JinaM}, a set of high-performance sentence embedding models trained on these datasets.
The \JinaM\ set is expected to comprise five distinct models, ranging in size from 35 million to 6 billion parameters.
Three of those models are already trained and published.~\footnote{\JinaS\ , \JinaB\ , \JinaL\ are available at \url{https://huggingface.co/jinaai}, and are also ranked in the MTEB leaderboard on Hugging Face: \url{https://huggingface.co/spaces/mteb/leaderboard}.}

The \JinaM\ models employ contrastive training on the T5 architecture~\citep{raffel2020exploring}.
It's important to note that we opt to use the T5 model as our base due to its pre-training on a mixed set of downstream tasks.
We argue that incorporating this approach can potentially enhance our ability to accurately gauge the effectiveness of our training strategy.

Our large-scale contrastive fine-tuning approach surpasses zero-shot T5 and delivers a performance level on par with other leading T5-based sentence embedding models such as Sentence-T5~\citep{ni2022sentence} and GTR~\citep{DBLP:journals/corr/abs-2112-07899}.
Consequently, this work demonstrates that high-quality sentence embeddings can be achieved with the judicious use of resources and innovative training methodologies.

\section{Dataset Preparation}
\label{sec:embedding_dataset}

In order to develop models that excel across a wide range of tasks, we collate a comprehensive set of both public and custom datasets.
These datasets target various retrieval objectives, such as e-commerce search, duplicate detection, web retrieval, article retrieval for question-answering, and text classification.
Consolidating these datasets into a unified format facilitates concurrent model training for all tasks.

\paragraph{Definition of Format:} 
Given the lack of non-relevance information in many of the datasets, we reformat each training item into pairs, designated as $(q, p) \in D_\mathit{pairs}$.
Each pair includes a query string $q$ and an associated target string $p$.
To leverage explicit non-relevance judgments, we create an auxiliary set of triplets $(q,p,n) \in D_\mathit{triplets}$, which pair a query string $q$ with a match $p$ (positive) and a non-matching string $n$ (negative).

\paragraph{Data Extraction:}
The methods used to extract pairs and triplets are specific to each source dataset.
For example, given a question-answer dataset, we use questions as query strings and answers as target strings.
Retrieval datasets often contain queries that can serve as query strings and relevant and non-relevant annotated documents which can operate as matching and non-matching strings.

\paragraph{Training Steps:}
Our training process is a two-step approach.
Initially, we train on pairs and then fine-tune the model using the triplets, as detailed in Section \ref{sec:train-triplets}.

\begin{figure*}[htb!]
\begin{minipage}[t]{0.5\textwidth}
\centering
\includegraphics[width=9cm]{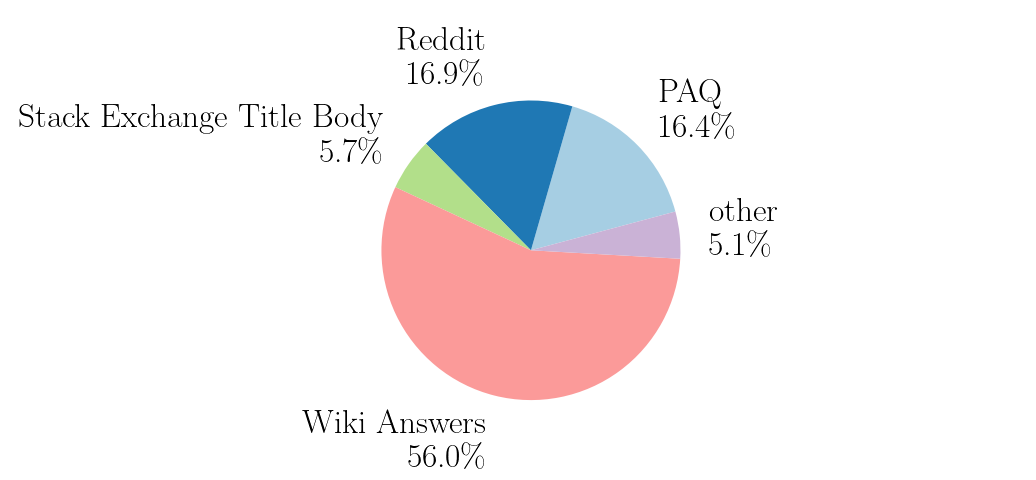}
(a) Original Distribution after Filtering
\end{minipage}
\begin{minipage}[t]{0.5\textwidth}
\centering
\includegraphics[width=9cm]{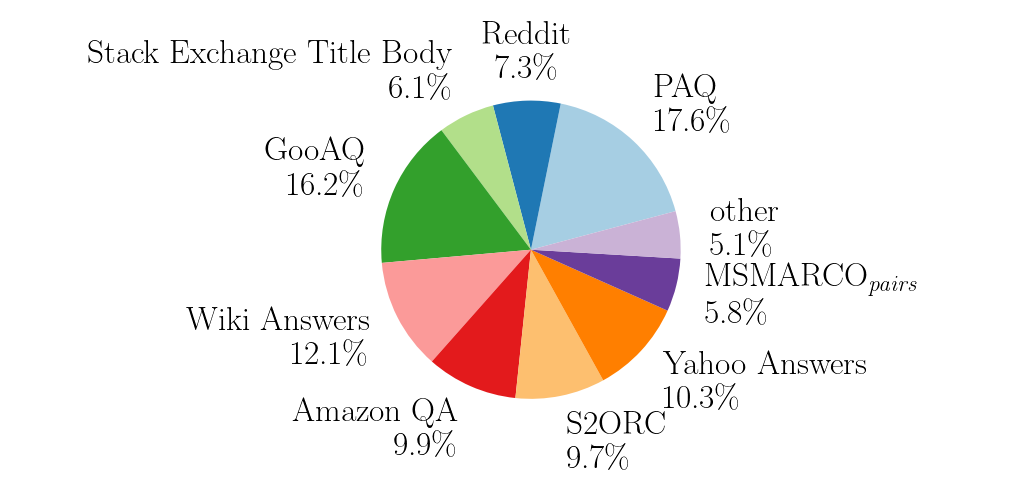}
(b) Adjusted by Sampling Rates
\end{minipage}
\caption{The composition of 385 million pairwise data}
\label{fig:sampling_rates}
\end{figure*}

\begin{figure}[htb!]
    \centering
    \includegraphics[width=7cm]{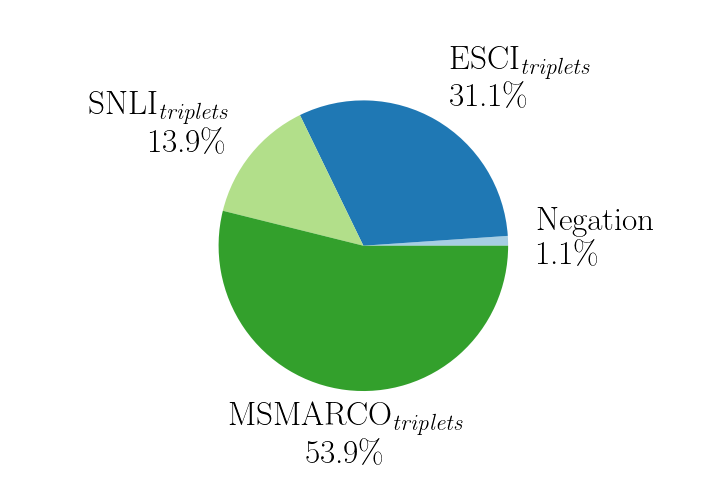}
    \caption{The composition of 927,000 triplets data}
    \label{fig:triplet-distribution}
\end{figure}

\subsection{Pairwise Data Preparation}

The substantial size and inconsistent quality of many large datasets necessitates a rigorous filtering pipeline.
We apply the following steps to filter training data:\
\\

\textbf{De-Duplication:}
Duplicated entries within training data can negatively impact model performance~\citep{hernandez2022scaling}, and potentially lead to overfitting.
Consequently, we remove duplicate entries from our dataset.
Considering the dataset's volume, we employ hash functions to identify and eliminate text pairs that map to duplicate hash values.
We normalize whitespace and capitalization before checking for duplicates.
Empty pairs and pairs with identical elements are also removed.\

\textbf{Language Filtering:}
Since we design our embedding models for English, we use the \texttt{fasttext-language-identification} model\footnote{\texttt{fasttext-language-identification} \tiny{(\url{https://huggingface.co/facebook/fasttext-language-identification})}} based on the fasttext text classification method~\citep{joulin2017bag} to remove non-English training items from the dataset.\

\textbf{Consistency Filtering:}
Consistency filtering means excluding training pairs with low semantic similarity.
Previous studies suggest that eliminating low-similarity pairs using an auxiliary, albeit less precise, model boosts performance~\citep{dai2023promptagator, wang2022text}.
We employ the \texttt{all-MiniLM-L6-v2} model\footnote{\texttt{all-MiniLM-L6-v2} model \tiny{(\url{https://huggingface.co/sentence-transformers/all-MiniLM-L6-v2})}} for consistency filtering in this manner: We generate embeddings for 1M pairs $(q_i,p_i)_i$ randomly sampled from $D_{pairs}$.
For every pair $(q,p) \in D_\mathit{pairs}$ in the dataset, we verify whether $p$ is among the top two passages most similar to $q$ based on the cosine similarity of their embeddings compared to all passages $p_i$, $i=1,...,1$M.\

The application of these preprocessing steps reduces the size of the dataset from over 1.5 billion mixed-quality pairs to 385 million high-quality pairs.
This reduction permits us to train our model with significantly less data than typical embedding models without sacrificing embedding quality.\footnote{For instance, models like \texttt{all-MiniLM-L6-v2} and \texttt{all-mpnet-base-v2} are trained on nearly 1.2 billion pairs, whereas other T5-based models such as \texttt{sentence-t5-base} or \texttt{sentence-t5-large} are trained on 2.2 billion pairs.}

\subsection{Triplet Data Preparation}

For the triplet dataset, we forego de-duplication and language filtering and we assume the quality of these datasets already meets our quality requirements.
However, we validate the relevance of the ``positive'' item with respect to the ``query'' for each triplet in a manner similar to consistency filtering.
Instead of contrasting the embedding cosine similarity $s(q,p)$ against a sample set, we compare it solely with the similarity $s(q,n)$ of the embeddings derived from the same triplet $(q,p,n) \in D_\mathit{triplets}$.
This is accomplished using a cross-encoder model, which evaluates the pair directly without generating embedding representations.
More specifically, we leverage the \texttt{ms-marco-MiniLM-L-6-v2} model\footnote{\texttt{ms-marco-MiniLM-L-6-v2} \tiny{(\url{https://huggingface.co/cross-encoder/ms-marco-MiniLM-L-6-v2})}} to verify whether the difference in retrieval scores determined by the model exceeds a threshold $r(q,p) - r(q,n) > \kappa$, with threshold $\kappa=0.2$, and eliminate all other pairs.
This methodology draws inspiration from the de-noising strategy proposed in~\citep{qu2021rocketqa}.

\subsection{Negation Data Preparation}

We observe that many embedding models struggle to accurately embed negations.
For instance, when embedding the three sentences: ``A couple walks hand in hand down a street.'', ``A couple is walking together.'', and ``A couple is not walking together.'', the first two should be embedded close together, while the second and third, contradictory in meaning, should be positioned further apart.\footnote{Although it could be argued that for certain tasks, like document retrieval, it might still be desirable for contradicting texts to be embedded closely. Regardless, in this example, the first two sentences should be assigned a higher similarity.}
However, for instance, the \texttt{all-MiniLM-L6-v2} model assigns a cosine similarity of 0.7 to the first two sentences, while attributing a similarity of 0.86 to the second and third.\footnote{Interestingly, our large model does \textit{correctly} assign a cosine similarity of 0.77 to the positive pair, and only a similarity of 0.62 to the negative pair, after fine-tuning with our negation dataset.}

We decide to address this problem by creating our own negation dataset\footnote{The negation dataset is available at {\url{https://huggingface.co/datasets/jinaai/negation-dataset}}}.
This dataset, based on positive pairs from the SNLI dataset\footnote{\url{https://huggingface.co/datasets/snli}} and negatives created with GPT-3.5, comprises triplets (anchor, entailment, negative) akin to the example given above, where (anchor, entailment) form a positive pair and the ``negative'' contradicts both the ``anchor'' and ``entailment'', while remaining syntactically very similar to ``entailment''.
This dataset forms a subset of our aforementioned triplet dataset, with training details provided in Section \ref{sec:train-triplets}.

Our model evaluation on the negation dataset, which includes a comparative analysis with other popular open-source models, is presented in Section \ref{sec:negation-evaluation}.

\subsection{Data Composition}

Our dataset of text pairs, represented as $D_\mathit{pairs} = D_1 \sqcup \dots \sqcup D_n$, is aggregated from 32 individual datasets.
This amounts to a total of 1.6 billion pairs before filtering, which is subsequently reduced to a robust 385 million high-quality pairs after rigorous filtering.

In comparison, our dataset of triplets initially comprises a total of 1.13 million entries before filtering, streamlined to 927,000 triplets after filtering.

The composition of our datasets after filtering is illustrated in Figure~\ref{fig:sampling_rates}a for the text pairs, and in Figure~\ref{fig:triplet-distribution} for the triplets.
Together, these form the final dataset for the training of the \JinaM\ models.

\section{Training}
\label{sec:training}

Training takes place in two distinct phases.
The first phase centers on training the model using the voluminous quantity of text pairs, consolidating the semantics of an entire text phrase into a single representative embedding.
The second phase uses the relatively small triplet dataset, comprising an anchor, an entailment, and a hard-negative, teaching it to differentiate between similar and dissimilar text phrases.

\subsection{Training on Pairwise Data}
\label{sec:train-pairs}

Each model within the \JinaM\ set is based on, and trained using, the zero-shot T5 models of corresponding size, as detailed in \citep{raffel2020exploring}.
The zero-shot T5 models are composed of encoder-decoder pairs.
However, \citet{ni2022sentence} has demonstrated that it is more effective to calculate text embeddings using only the encoder component of the T5 models, as opposed to deploying both encoder and decoder.
Consequently, the \JinaM\ models use only the encoders of their respective T5 models. 

During tokenization, \JinaM\ models use SentencePiece~\citep{kudo2018sentencepiece} to segment input text and encode them into WordPiece tokens~\citep{kudo2018subword}.
Following the encoder model, a mean pooling layer is implemented to generate fixed-length representations from the token embeddings.

For the training process involving pairs, we employ InfoNCE~\citep{DBLP:journals/corr/abs-1807-03748}, a contrastive loss function.
This function calculates the loss for a pair $(q,p) \sim B$ within a batch $B \in D^k$ of text pairs, where the batch size is $k$, as follows:

\begin{align*} 
    \mathcal{L}_{\nce}^{pairs}(B) := \mathbb{E}_{(q,p)\sim B}\Bigg[-\ln \frac{e^{s(q, p)/\tau}}{\sum_{i = 1}^k e^{s(q, p_i)/ \tau}}\Bigg]
\end{align*}
The loss is calculated by comparing the cosine similarity between a given question $q$ and its target $p$, with the similarity to all other targets in the batch.
% Typically when calculating on data formatted as pairs $[(q_1, k_1), (q_2, k_2), ... (q_K, k_K)$], the loss is calculated by comparing the similarity between and $q_i$ and $[k_1, ... k_K]$, however 
We found that calculating the loss in both directions results in greater improvements during training.
Accordingly,  the loss is defined as follows:

\begin{align*}
    \mathcal{L}^{pairs}(B) &:= \mathcal{L}^{pairs}_{\nce}(B) + \mathcal{L}^{pairs}_{\overline{\nce}}(B),\text{ where} \\
    \mathcal{L}_{\overline{\nce}}^{pairs}(B) &:= \mathbb{E}_{(q,p)\sim B}\Bigg[-\ln \frac{e^{s(p, q) / \tau}}{\sum_{i = 1}^k e^{s(p, q_i) / \tau}}\Bigg]. 
\end{align*}

Intuitively, $\mathcal{L}^{pairs}_{\overline{\nce}}$ matches the target string to all query strings instead.
The constant $\tau$ denotes a temperature parameter which we set to $\tau=0.05$.
This method of calculating the loss is based on a similar method in \citep{DBLP:journals/corr/abs-2201-10005}.

\subsection{Data Sampling in Pairwise Training}
\label{subsec:sampling}

Rather than sequentially training on individual datasets, we opt for a parallel approach, training on all datasets concurrently.
We postulate that this parallel training promotes enhanced model generalization across diverse tasks. Despite this, each training batch is exclusively composed of data from a single dataset. 
This ensures that loss calculations, performed across the entire batch, do not conflate data from different tasks. 

Our dataloader operates by initially selecting a dataset, followed by sampling the requisite number of data points from it to constitute a batch for the worker (refer to Section~\ref{sec:training-setup}).
Prior to training, the pairs within the datasets are thoroughly shuffled.

Sampling a dataset $D_i$ follows a probability distribution $\rho$ across all datasets $D_i$.
The probability of sampling $D_i$ is $\rho\left(D_i\right)=\frac{|D_i|s_i}{\sum_{j=1}^n |D_j|s_j}$ and is contingent upon the dataset's size $|D_i|$ and a scaling factor $s_i$.

Given the disparity in dataset sizes, it is critical to frequently sample from larger datasets to prevent overfitting on the smaller ones.
Furthermore, we manipulate the sampling rates of datasets using scaling factors to prioritize training on high-quality datasets and achieve balance among text domains.
In scenarios where datasets with higher sampling rates deplete their items before the completion of a training epoch, the dataset is reset, enabling the model to cycle through its items anew.
This ensures that high-sampling-rate datasets contribute multiple times within a single training epoch.

Figure~\ref{fig:sampling_rates}b displays the proportion of each dataset used based on their sampling rates. Following the creation of this adjusted distribution, the frequency of sampling from larger datasets significantly diminishes, resulting in only 180 million pairs actually being used during training.

\subsection{Training on Triplet Data}
\label{sec:train-triplets}

Following the completion of pairwise training, the model progresses to the next phase which involves training on the triplet datasets.
This phase uses a different loss function, leveraging negatives for improved model performance.

We experimented with various triplet loss functions and found that the best results are achieved through a combination of multiple commonly used triplet loss functions.
Specifically, we use the extended version of the InfoNCE loss $\mathcal{L}^{triplets}_\mathit{NCE+}$, given by \eqref{eq:triplet-loss-negatives}, which employs additional negatives~\citep{reimers2023mnrl}, the reverse InfoNCE loss $\mathcal{L}^{triplets}_{\overline{\nce}}$ from the initial training phase as given by \eqref{eq:triplet-loss-reverse}, and the triplet margin loss function $\mathcal{L}^{triplets}_3$ as presented in \eqref{eq:triplet-loss-margin}~\citep{chechik2010large}.

The triplet function $\mathcal{L}^{triplets}_3$ determines the cosine similarity difference between the query and target $s(q,n)$, and the query and negative match $s(q,n)$.
Furthermore, it establishes a minimal margin $\varepsilon = 0.05$ between these two values.
If the negative is more similar to the query or the margin is violated, $\mathcal{L}^{triplets}_3$ returns a positive value.
Otherwise, it yields $0$, which is achieved through the application of the $\text{ReLU}$ activation function.
For the temperature parameter, we opted for a value of $\tau=0.05$.

% \newpage % this align needs manual formatting, hence the newpage
\begin{figure*}
    
% \begin{strip}
        \begin{align}
            \mathcal{L}^{triplets}(B) &:= \mathcal{L}^{triplets}_\mathit{NCE+}(B) + \mathcal{L}_{\overline{\nce}}^{triplets}(B) + \mathcal{L}^{triplets}_3(B) \label{eq:triplet-loss-mean} ,\quad\text{where}&&  \\
            \mathcal{L}^{triplets}_\mathit{NCE+}(B) &:= \mathbb{E}_{(q,p,n)\sim B}\Bigg[-\ln \frac{\exp(s(q, p) / \tau)}{\sum_{i = 1}^k \exp(s(q, p_i) / \tau)+ \exp(s(q, n_i) / \tau)}\Bigg] \label{eq:triplet-loss-negatives}, && \\
            \mathcal{L}^{triplets}_{\overline{\nce}}(B) &:= \mathbb{E}_{(q,p,n)\sim B}\Bigg[-\ln \frac{\exp(s(p, q) / \tau)}{\sum_{i = 1}^k \exp(s(p, q_i) / \tau)}\Bigg] \label{eq:triplet-loss-reverse}, && \\
            \mathcal{L}^{triplets}_3(B) &:= \mathbb{E}_{(q,p,n)\sim B}\Bigg[\text{ReLU}\Big(s(q,n)-s(q,p) + \varepsilon\Big)\Bigg] \label{eq:triplet-loss-margin}. &&
        \end{align}
% \end{strip}
\end{figure*}

\section{Evaluation}
\label{sec:training-setup}

We conduct a comprehensive evaluation to compare our models against other state-of-the-art models (Section~\ref{sec:evaluation}), investigate the impact of our filtering pipeline (Section~\ref{sec:ablation_study}), and evaluate the models' sensitivity to negation of statements (Section~\ref{sec:negation-evaluation}).
Section~\ref{sce:train_details} mentions details about the training.

To provide comprehensive results on the performance of models on various downstream tasks applicable to embeddings, we rely on the MTEB benchmark frameworks introduced by~\citet{muennighoff2023mteb}.
This also compromises all the retrieval tasks included in the BEIR~\cite{thakur2021beir} benchmark.
We also publish the code for executing it on our models on the Hugging Face pages of our model\footnote{\url{https://huggingface.co/jinaai/jina-embedding-b-en-v1/blob/main/mteb_evaluation.py}}.
For evaluating models on the negation dataset, we use our own separate evaluation tool\footnote{\url{https://huggingface.co/jinaai/jina-embedding-b-en-v1/blob/main/negation_evaluation.py}}.

\subsection{Performance Against State-of-the-Art Models}
\label{sec:evaluation}

\begin{table}[htb!]
\begin{tabular}{lrr}
\toprule
Model                  & Parameters & \makecell{Embedding\\Dimensions} \\
\midrule
sentence-t5-xxl        & 4.9b                                  & 768                                       \\
gtr-t5-xxl             & 4.9b                                  & 768                                       \\
gtr-t5-xl              & 1.2b                                  & 768                                       \\
sentence-t5-xl         & 1.2b                                  & 768                                       \\ \midrule
\JinaL             & 330m                                  & 1024                                      \\
gtr-t5-large           & 330m                                  & 768                                       \\
sentence-t5-large      & 330m                                  & 768                                       \\ \midrule
all-mpnet-base-v2      & 110m                                  & 768                                       \\
\JinaB             & 110m                                  & 768                                       \\
gtr-t5-base            & 110m                                  & 768                                       \\
sentence-t5-base       & 110m                                  & 768                                       \\ 
\midrule
\JinaS             & 35m                                  & 512                                       \\ 
all-MiniLM-L6-v2       & 23m                                  & 384                                       \\
\bottomrule
\end{tabular}
\caption{Model sizes and output dimensions}
\label{tab:model-sizes}
\end{table}

To gauge the performance of the \JinaM\ set in relation to other similarly sized open-source and close-sourced models, we select representative models from five distinct size categories, as depicted in Table \ref{tab:model-sizes}.
Additionally, we include sentence-t5 and gtr-t5 xl and xxl models, which are based on T5 models with 3 billion and 11 billion parameters, respectively.
This inclusion allows investigating the performance variation with models of such massive scales.

Table \ref{tab:STS-scores} presents the scores for MTEB's sentence similarity tasks, wherein the models within the \JinaM\ set outshine their similarly sized counterparts across numerous tasks.
Notably, the \JinaL\ model consistently delivers comparable, if not superior, results to models in the billion-parameter scale.
\JinaB\ and \JinaS\ also exhibit competitive performances with models of analogous sizes, exceeding their peers on the BIOSSES\footnote{\url{https://tabilab.cmpe.boun.edu.tr/BIOSSES/DataSet.html}} task.
This highlights the benefits of training with highly diverse data sources.

\JinaB\ consistently demonstrates performances similar to or better than gtr-t5-base, which was trained specifically for retrieval tasks \cite{DBLP:journals/corr/abs-2112-07899}.
However, it seldom matches the scores of sentence-t5-base, which was trained on sentence similarity tasks \cite{ni2022sentence}.

The evaluation of model performances on retrieval tasks, presented in Table \ref{tab:retrieval-scores}, reflects a similar relationship among gtr-t5, sentence-t5, and \JinaM.
Here, gtr-t5 models, which have been specially trained on retrieval tasks, consistently score the highest for their respective sizes.
\JinaM\ models follow closely behind, whereas sentence-t5 models trail significantly.
The \JinaM\ set's capability to maintain competitive scores across these tasks underscores the advantage of multi-task training.

As illustrated in Table \ref{tab:reranking-scores}, \JinaL\ also achieves exceedingly high scores on reranking tasks, often outperforming larger models.
Similarly, \JinaB\ surpasses gtr-t5-large and sentence-t5-large on several reranking tasks, which could once again be attributed to the specific training tasks of sentence-t5 and gtr-t5.

\begin{table}[htb!]
\begin{tabular}{lccc}
\toprule
Model        & RR & RT & STS \\ 
\midrule
sentence-t5-xxl        & 56.42                                              & 42.24                                               & \textbf{82.63}                                 \\
gtr-t5-xxl             & \textbf{56.66}                                              & \textbf{48.48}                                               & 78.38                                          \\
gtr-t5-xl              & 55.96                                              & 47.96                                               & 77.80                                           \\
sentence-t5-xl         & 54.71                                              & 38.47                                               & 81.66                                          \\ \midrule
\JinaL & \textbf{56.42}                                    & 44.81                                & 80.96                                          \\
gtr-t5-large           & 55.36                                              & \textbf{47.42}                                      & 78.19                                          \\
sentence-t5-large      & 54.00                                                 & 36.71                                               & \textbf{81.83}                                          \\ \midrule
all-mpnet-base-v2      & \textbf{59.36}                                              & 43.81                                               & 80.28                                          \\
\JinaB             & 55.84                                              & 44.03                                & 79.93                                          \\
gtr-t5-base            & 54.23                                              & \textbf{44.67}                                               & 77.07                                          \\
sentence-t5-base       & 53.09                                              & 33.63                                               & \textbf{81.14}                                          \\ \midrule
\JinaS & 53.07                                              & 38.91                                               & 78.06                                          \\
all-MiniLM-L6-v2       & \textbf{58.04}                                              & \textbf{41.95}                                               & \textbf{78.90}                                           \\ \bottomrule
\end{tabular}
\caption{Average Scores for Reranking (RR), Retrieval (RT) and sentence similarity tasks (STS)}
\label{model-scores}
\end{table}

\subsection{Impact of Filtering Steps}
\label{sec:ablation_study}

We evaluate the effectiveness of our dataset preprocessing pipeline by performing an ablation study.
In this study, we fine-tune our smallest model on the Reddit dataset, where various preprocessing steps are individually applied.
The corresponding results are presented in Table~\ref{tab:preprocessing_evaluation}.

\begin{table*}[htb]
    \centering
\begin{tabularx}{\linewidth}{lX}
  \toprule
  \multicolumn{2}{c}{Retrieval} \\
  Data Preparation & \hfill Quora \hfill SciFact \hfill Trec-Cov \hfill\null \\
  \midrule
  No Extra Filter & \hfill 0.734 \hfill 0.218 \hfill 0.242 \hfill\null \\
  Language & \hfill 0.741 \hfill 0.218 \hfill 0.250  \hfill\null \\
  Consistency & \hfill 0.805 \hfill \bf{0.381} \hfill \textnormal{0.297}  \hfill\null \\
  Language + Consistency & \hfill \bf{0.806} \hfill \textnormal{0.379} \hfill \bf{0.306} \hfill\null \\
  \bottomrule
  \toprule
  \multicolumn{2}{c}{Sentence Similarity} \\
  Data Preparation & STS12 \hfill STS13 \hfill STS14 \hfill STS15 \hfill STS16 \hfill STS17 \hfill STS22 \hfill\null \\
  \midrule
  No Extra Filter & 0.558 \hfill 0.668 \hfill 0.573 \hfill 0.694 \hfill 0.706 \hfill 0.764  \hfill 0.606 \hfill\null \\
  Language        & 0.561  \hfill 0.668 \hfill 0.579 \hfill 0.697 \hfill 0.704 \hfill 0.765 \hfill 0.609 \hfill\null \\
  Consistency     & 0.652 \hfill \bf{0.728} \hfill \textnormal{0.652} \hfill \textnormal{0.760} \hfill \textnormal{0.755} \hfill \textnormal{0.808} \hfill \bf{0.610} \hfill\null \\
  Language + Consistency & \bf{0.653} \hfill \textnormal{0.727} \hfill \bf{0.656} \hfill \bf{0.764} \hfill \bf{0.757} \hfill \bf{0.810} \hfill \textnormal{0.609} \hfill\null \\
  \bottomrule
\end{tabularx}
    \caption{Evaluation of Data-Preparation Effectiveness on the Reddit Dataset.
             Retrieval evaluated on nDCG@10, Sentence Similarity on Spearman.}
    \label{tab:preprocessing_evaluation}
\end{table*}

The ablation study's results underscore the value of both language and consistency filtering as crucial preprocessing steps.
Their combined application results in the highest performance across the majority of benchmarks. 

Specifically for the Reddit dataset, we observe a significant performance boost with the application of consistency filtering, while language filtering only marginally enhances the performance.
We can account for this disparity by noting that the language filter removes only 17.4\% of the Reddit data, while consistency filtering screens out 84.3\footnote{It is pertinent to note that the subsets filtered out overlap, thus the combined application of language and consistency filtering filters out \textit{only} 86.8\% of the data.}. Reddit samples are primarily in English, but many are positive pairs with very low similarity, making consistency filtering more effective than language filtering.

The effectiveness of these preprocessing steps, however, does exhibit variability across different datasets.

\subsection{Effectiveness of Negation Data}
\label{sec:negation-evaluation}

To determine the effectiveness of our models on negation data, we evaluate them against the test split of our negation dataset, comparing the results with other open source models. 
We measure performance with respect to two metrics: one measures the percentage of samples where the model positions the \textit{anchor} and \textit{entailment} closer than the \textit{anchor} and \textit{negative} (which is an easy task, as the \textit{anchor} and \textit{negative} are syntactically dissimilar), the other measures the percentage of samples where the model positions the \textit{anchor} and \textit{entailment} closer than the \textit{entailment} and \textit{negative} (which is a hard task, as the \textit{entailment} and \textit{negative} are syntactically more similar than the \textit{anchor} and \textit{entailment}).
The former is denoted by \textit{EasyNegation}, the latter by \textit{HardNegation}.
The outcomes of these evaluations are displayed in Table~\ref{tab:negation_evaluation}.
We assess our models both before and after fine-tuning on the triplet data, denoted as \textbf{<model>\tiny{pairwise}} and \textbf{<model>\tiny{all}}, respectively.

From the results, we observe that across all model sizes, fine-tuning on triplet data (which includes our negation training dataset) dramatically enhances performance, particularly on the HardNegation task.
Our models are on par with other state-of-the-art open-source models in terms of performance, while achieving this with only a fraction of the training data required by their counterparts.

\begin{table*}[htb]
    \centering
\begin{tabular}{lrrrr}
  \toprule
  & EasyNegation & HardNegation & Parameters & Training samples\\
  \midrule
  \JinaS \tiny{pairwise} & 88.4\% & 8.4\% & 35m & 385m\\
  \JinaB \tiny{pairwise} & 93.0\% & 13.8\% & 110m & 385m \\
  \JinaL \tiny{pairwise} & 94.6\% & 16.6\% & 330m & 385m\\
  \JinaS \tiny{all} & 96.6\% & 35.2\% & 35m & 386m \\
  \JinaB \tiny{all} & 97.8\% & 54.6\% & 110m & 386m \\
  \JinaL \tiny{all} & \textbf{98.2\%} & 65.4\% & 330m & 386m \\
  \midrule
  all-MiniLM-L6-v2 & 94.8\% & 29.4\% & 23m & 1170m\\
  all-mpnet-base-v2 & 97.4\% & \textbf{67.6\%} & 110m & 1170m\\
  sentence-t5-base & 96.0\% & 55\% & 110m & 2275m\\
  sentence-t5-large & \textbf{98.2\%} & 64.0\% & 330m & 2275m\\
  \bottomrule
\end{tabular}
    \caption{Evaluating a Range of Models on the Negation Dataset: A Benchmark Analysis of \JinaM\ Trained on Both Pairwise-Only and Combined Pairwise and Triplet Data. The negation dataset is available at {\url{https://huggingface.co/datasets/jinaai/negation-dataset}}}
    \label{tab:negation_evaluation}
\end{table*}

\section{Related Work}
\label{sec:related_work}

The field of embedding models has seen significant advanced over the years, with the development of various models featuring diverse architectures and training pipelines.
For instance, Sentence-BERT \cite{reimers2019sentence} uses BERT to generate sentence embeddings. Similarly, Sentence-T5 \cite{ni2022sentence}, based on the encoder architecture of T5, demonstrates superior performance over Sentence-BERT on numerous benchmarks.
The study underscores the effectiveness of encoders for sentence embeddings, contrasting with another approach that explores the use of decoders \cite{muennighoff2022sgpt}.

Knowledge distillation \cite{hinton2015distilling} offers an alternative approach to model training.
In this setup, a larger, pre-trained model acts as a mentor, instructing a smaller model during training.
This methodology can be seamlessly integrated with a contrastive loss function, presenting an avenue for future investigation.

Embedding models can also be characterized based on their functionality. For instance, while some models are designed to solely embed queries, others are trained to embed queries along with specific instructions, generating task-dependent embeddings \cite{su2023embedder}.
An example of this using a T5-based model is the large dual encoder \cite{DBLP:journals/corr/abs-2112-07899}, which is fine-tuned for retrieval tasks and computes a retrieval score directly.

Recent studies \cite{DBLP:journals/corr/abs-2201-10005, wang2022text} emphasize the benefits of contrastive pre-training coupled with fine-tuning on hard negatives.
Both approaches have achieve state-of-the-art results on multiple benchmarks, with \cite{wang2022text} also employing consistency filtering as part of their preprocessing pipeline.

% As part of our evaluation process, we utilized two widely-accepted benchmarks for embedding models: MTEB \cite{muennighoff2023mteb} and BEIR \cite{thakur2021beir}.

\section{Training Details}
\label{sce:train_details}

\begin{table}[t]
    \centering
    \begin{tabular}{lr}
    \toprule
    Hyperparameters & Value   \\
    \midrule
    \# of devices   & 8       \\
    Sequence length & 512     \\
    Model precision & 32 bit     \\
    Learning rate   & 0.00005 \\
    \# of steps for learning rate warm-up & 500     \\
    Batch size for \JinaS & 4096 \\
    Batch size for \JinaB & 2048 \\
    Batch size for \JinaL & 1024 \\
    \bottomrule
    \end{tabular}
    \caption{Hyperparameters}
    \label{tab:hparams}
\end{table}

For training, we employ A100 GPUs and leverage the DeepSpeed stage 2 distributed training strategy \citep{rajbhandari2020zero} for effective multi-device management.
For training our models we use the AdamW optimizer, coupled with a learning rate scheduler that adjusts the learning rate during the initial stages of training.
The hyperparameters used across all three models throughout the training process are listed in Table \ref{tab:hparams}.

\section{Conclusion}
\label{sec:conclusion}

This paper introduces the \JinaM\ set of embedding models, demonstrating that competitive performance on various tasks can be achieved while substantially reducing the amount of training data, when compared to other models with comparable backbones.
Through an extensive evaluation on the MTEB benchmark, we show that employing judicious data filtering techniques can lead to enhanced performance in comparison to training with a larger, yet lower-quality dataset.
These findings significantly shift the paradigm, indicating that training large language models for embedding tasks can be conducted with less data than previously assumed, leading to potential savings in training time and resources.

However, we acknowledge the limitations of the current methodologies and the performance of the \JinaM\ set.
During the training on pairs, the sampling rate selection was based on a heuristic approach.
Given the vast size of the search space for these sampling rates, we leaned on our intuition and dataset familiarity to prioritize higher-value datasets over their lower-value counterparts.
This subjective approach, however, points to the need for more objective methods for future advancements. 

Additionally, the \JinaM\ set fell short on some tasks.
For instance, calculating sentence similarity on our negation dataset (as described in Section \ref{sec:negation-evaluation}) didn't meet our expectations (see Table~\ref{tab:negation_evaluation}) nor achieves competitive scores for classification and clustering tasks on the MTEB benchmark. These performance shortcomings suggest a possible deficit in the representation of these types of tasks in our training data, necessitating further investigation.

Looking ahead, we aim to refine our training processes to deliver models with improved performance and greater sequence length.
Our future endeavors also include generating bilingual training data and training an embedding model capable of understanding and translating between two languages, thereby expanding the utility and versatility of the \JinaM\ set.

\balance

\bibliographystyle{unsrtnat}
\bibliography{references}  %%% Uncomment this line and comment out the ``thebibliography'' section below to use the external .bib file (using bibtex) .

\clearpage
\pagenumbering{gobble}
\onecolumn
\appendix

\section*{Appendix}

\begin{table*}[htb!]
%\hspace{-2cm}
\centering
\small{
\setlength\tabcolsep{5pt}
\begin{tabular}{|l|cccccccccc|}
\hline
Model                  & \multicolumn{1}{l}{BIOSSES} & \multicolumn{1}{l}{SICK-R} & \multicolumn{1}{l}{STS12} & \multicolumn{1}{l}{STS13} & \multicolumn{1}{l}{STS14} & \multicolumn{1}{l}{STS15} & \multicolumn{1}{l}{STS16} & \multicolumn{1}{l}{STS17} & \multicolumn{1}{l}{STS22} & \multicolumn{1}{l|}{\makecell{STS\\Benchm.}} \\ \hline
sentence-t5-xxl        & 80.43                       & \textbf{80.47}             & 78.85                     & \textbf{88.94}            & \textbf{84.86}            & \textbf{89.32}            & 84.67                     & \textbf{89.46}                    & 65.33                          & \textbf{84.01}                    \\
sentence-t5-xl         & 73.12                       & 79.98                      & \textbf{79.02}            & 88.80                      & 84.33                     & 88.89                     & \textbf{85.31}            & 88.91                             & 64.32                          & 83.93                             \\
gtr-t5-xxl             & \textbf{81.91}              & 74.29                      & 70.12                     & 82.72                     & 78.24                     & 86.26                     & 81.61                     & 85.18                             & 65.76                          & 77.73                             \\
gtr-t5-xl              & 78.94                       & 73.63                      & 69.11                     & 81.82                     & 77.07                     & 86.01                     & 82.23                     & 84.90                              & \textbf{66.61}                 & 77.65                             \\ \hline
sentence-t5-large      & 78.93                       & \textbf{80.34}             & \textbf{79.11}            & \textbf{87.33}            & \textbf{83.17}            & \textbf{88.28}            & \textbf{84.36}            & 88.99                             & 62.39                          & \textbf{85.36}                    \\
gtr-t5-large           & \textbf{84.86}              & 73.39                      & 70.33                     & 82.19                     & 77.16                     & 86.31                     & 81.85                     & 83.93                             & 64.30                           & 77.60                              \\
\JinaL & 84.43                       & 79.20                      & 74.53                     & 83.16                    & 78.09                     & 86.91                     & 83.65                     & \textbf{90.16}                    & \textbf{64.89}                 & 84.60                             \\
 \hline
 sentence-t5-base       & 75.89                       & 80.18                      & \textbf{78.05}            & \textbf{85.85}            & \textbf{82.19}            & \textbf{87.46}            & \textbf{84.03}            & \textbf{89.57}                    & 62.66                          & \textbf{85.52}                    \\
gtr-t5-base            & 79.00                          & 71.45                      & 68.59                     & 79.09                     & 74.64                     & 84.85                     & 81.57                     & 85.80                              & 66.17                          & 79.58                             \\
all-mpnet-base-v2      & 80.43                       & \textbf{80.59}             & 72.63                     & 83.48                     & 78.00                        & 85.66                     & 80.03                     & 90.60                              & \textbf{67.95}                 & 83.42                             \\
\JinaB & \textbf{83.58}              & 79.14                      & 75.06                     & 80.86                     & 76.13                     & 85.55                     & 81.21                     & 88.98                             & 66.22                          & 82.57                             \\ \hline
all-MiniLM-L6-v2       & 81.64                       & \textbf{77.58}             & 72.37                     & \textbf{80.60}             & \textbf{75.59}            & \textbf{85.39}            & 78.99                     & \textbf{87.59}                    & \textbf{67.21}                 & \textbf{82.03}                    \\
\JinaS & \textbf{82.96}              & 76.33                      & \textbf{74.28}            & 78.55                     & 73.84                     & 83.71                     & \textbf{80.03}            & 87.49                             & 64.25                          & 79.20                              \\ \hline
text-emb-ada-002* & 86.35              & 80.60                      & 69.80            & 83.27                     & 76.09                     & 86.12                     & 85.96            & 90.25                             & 68.12                          & 83.17                              \\ \hline
\end{tabular}
}
\caption{Spearman Correlation for Sentence Similarity Tasks}
\label{tab:STS-scores}
\end{table*}
\let\thefootnote\relax\footnotetext{* text-emb-ada-002 appears in a separate category since no model size is known and the embedding size is much higher compared to other models.}
\begin{table*}[h!]
\centering
\small{
\begin{tabular}{|l|cccc|}
\hline
Model                        & \makecell{AskUbuntu- \\ DupQuestions} & \makecell{MindSmall- \\ Reranking} & SciDocsRR & \makecell{StackOverflow- \\ DupQuestions} \\ \hline
sentence-t5-xxl              & \textbf{66.16}                            & 30.60                                   & 76.09                         & 52.85                                          \\
sentence-t5-xl         & 62.86                                     & 29.77                                  & 75.16                         & 51.05                                          \\
gtr-t5-xxl                   & 63.23                                     & \textbf{31.93}                         & \textbf{77.96}                & \textbf{53.50}                                  \\
gtr-t5-xl                    & 63.08                                     & 31.50                                   & 76.49                         & 52.79                                          \\ \hline
sentence-t5-large      & 61.51                                     & 30.27                                  & 74.88                         & 49.34                                          \\
gtr-t5-large           & 61.64                                     & \textbf{31.84}                         & 76.39                         & \textbf{51.58}                                 \\
\JinaL & \textbf{62.83}                            & 31.48                                  & \textbf{80.97}                & 50.38                                          \\ \hline
sentence-t5-base       & 59.73                                     & 30.20                                   & 73.96                         & 48.46                                          \\ 
gtr-t5-base            & 60.86                                     & 31.33                         & 73.71                         & 51.01                                          \\
all-mpnet-base-v2      & \textbf{65.85}                            & 30.97                                  & \textbf{88.65}                & \textbf{51.98}                                 \\
\JinaB & 62.40                                     & \textbf{31.56}                                  & 79.31                         & 50.11                                          \\ \hline
all-MiniLM-L6-v2       & \textbf{63.48}                            & \textbf{30.80}                          & \textbf{87.12}                & \textbf{50.76}                                 \\
\JinaS & 60.25                                     & 30.68                                  & 74.16                         & 47.18                                          \\ \hline
text-emb-ada-002* & 62.05                                     & 31.45                                  & 81.22                         & 50.54                                          \\ \hline
\end{tabular}
}
\caption{Mean Average Precision (mAP@10) for Reranking Tasks}
\label{tab:reranking-scores}
\end{table*}

\begin{landscape}
\begin{table*}[h!]
\centering
\small{
\begin{tabular}{|l|cccccccccccccc|}
\hline
Model                  & FEVER & HotpotQA & MSMARCO & NQ & \makecell{Quora\\ Retrieval} & SciFact & \makecell{TREC\\ COVID} & \makecell{Argu\\ Ana} & \makecell{Climate\\ FEVER} & DBPedia & \makecell{FiQA\\ 2018} & NFCorpus & SCIDOCS & \makecell{Touche\\ 2020}\\ \hline
sentence-t5-xxl        & 51.20                     & 42.14                        & 27.67                       & 52.87                  & 85.96                              & 55.38                       & 59.48                & 39.85            & 14.63                  & 39.19                       & 46.68                        & \textbf{35.08 }                       & \textbf{17.17}                       & 21.65                      \\
sentence-t5-xl         & 36.12                     & 37.17                        & 25.17                       & 46.29                  & 85.85                              & 50.91                       & 54.77            & 39.40                       & 10.61                            & 33.65                       & 44.71                        & 33.18                        & 15.97                       & 22.51                            \\
gtr-t5-xxl             & \textbf{74.08}                     & \textbf{59.67}                        & \textbf{44.05}                       & \textbf{57.24}                  & \textbf{89.09}                              & \textbf{66.77}                       & 51.90              & \textbf{53.77}                       & \textbf{27.21}                            & \textbf{41.28}              & \textbf{46.78 }                       & 34.18                        & 15.88                       & \textbf{26.76}                  \\
gtr-t5-xl              & 72.18                     & 58.91                        & 43.52                       & 56.16                  & 88.91                              & 64.2                        & \textbf{60.09}                  & 52.81                       & 27.01                            & 39.74                       & 44.19                        & 33.34                        & 15.71                       & 25.26                       \\ \hline

sentence-t5-large      & 36.21                     & 33.95                        & 23.96                       & 42.02                  & 85.73                              & 49.91                       & 46.11              & 39.27                       & 11.36                            & 31.55                       & \textbf{43.55}                        & 31.10                        & 15.38                       & 21.63                        \\ 
gtr-t5-large           & \textbf{72.66}                     & \textbf{57.85}                        & \textbf{42.73}                       & \textbf{55.09}                  & \textbf{88.47}                              & \textbf{63.42}                       & 56.68             & \textbf{52.09}              & \textbf{26.90}                            & \textbf{39.55}                       & 42.79                        & \textbf{32.63}                        & 15.51                       & \textbf{28.29}                                 \\
\JinaL & 71.90                     & 54.95                        & 40.34                       & 51.40                  & 88.09                              & 59.76                       & \textbf{57.25}       & 46.48                       & 21.26                   & 34.13                       & 37.27                        & 32.24                        & \textbf{18.45}                       & 20.73                               \\ \hline
sentence-t5-base       & 26.17                     & 33.20                        & 20.70                       & 36.32                  & 85.49                              & 45.76                       & 40.70            & 44.85                       & 10.37                            & 27.77                       & 34.83                        & 28.65                        & 14.15                       & 20.30                         \\
gtr-t5-base            & 68.93                     & \textbf{54.93}                        & \textbf{41.16}                       & \textbf{50.47}                  & \textbf{87.98}                              & 59.74                       & 56.05               & \textbf{50.83}                       & \textbf{24.88}                            & \textbf{35.24}                       & 35.15                        & 30.22                        & 14.00                       & \textbf{25.89}                             \\
all-mpnet-base-v2      & 50.86                     & 39.29                        & 39.75                       & 50.45                  & 87.46                              & \textbf{65.57}                       & 51.33            & 46.52                       & 21.97                            & 32.09                       & \textbf{49.96}                        & \textbf{33.29 }                       & \textbf{23.76}                      & 19.93                                 \\
\JinaB & \textbf{73.29}                     & 52.78                        & 37.77                       & 47.87                  & 87.63                              & 59.40                       & \textbf{60.57}              & 49.01                       & 21.48                            & 32.44                       & 34.06                        & 30.38                        & 17.63                       & 18.59                 \\ \hline
all-MiniLM-L6-v2       & 51.93                     & 46.51                        & \textbf{36.54}                       & \textbf{43.87}                  & \textbf{87.56}                              & \textbf{64.51}                       & 47.25                  & \textbf{50.17}                       & \textbf{20.27}                            & \textbf{32.33}                       & \textbf{36.87}                        & \textbf{31.59}                        & \textbf{21.64}                       & \textbf{16.90}                        \\ 
\JinaS & \textbf{69.12}                     & \textbf{47.48}                        & 31.80                       & 38.89                  & 85.69                              & 52.40                       & \textbf{52.30}             & 43.57                       & 17.25                            & 28.28                       & 25.19                        & 25.96                        & 15.29                       & 16.67                     \\ \hline
text-emb-ada-002* & 74.99                     & 60.90                        & 40.91                       & 51.58                  & 87.60                              & 72.75                       & 68.47             & 57.44                       & 21.64                            & 39.39                       & 44.41                        & 36.97                        & 18.36                       & 21.61                     \\ \hline
\end{tabular}
}
\caption{Normalized Discounted Cumulative Gain (nDCG@10) for retrieval tasks}
\label{tab:retrieval-scores}
\end{table*}
\end{landscape}





% \end{thebibliography}

\end{document}